# Evaluation of the Automated Labeling Method for Taxonomic Nomenclature Through Prompt-Optimized Large Language Model

Keito Inoshita
*Faculty of Business and Commerce*
*Kansai University*
*Osaka, Japan*
*Email: k845578@kansai-u.ac.jp*

Kota Nojiri
*Graduate School of Agricultural and Life Sciences,*
*The University Museum*
*The University of Tokyo*
*Bunkyo, Japan*
*Email: kotanojiri@g.ecc.u-tokyo.ac.jp*

Haruto Sugeno
*Faculty of Symbiotic Systems Science*
*Fukushima University*
*Fukushima, Japan*
*Email: s2470053@ipc.fukushima-u.ac.jp*

Takumi Taga
*Graduate School of Pharmaceutical Sciences*
*Nagoya University*
*Nagoya, Japan*
*Email: taga.takumi.h9@s.mail.nagoya-u.ac.jp*

*Abstract*—**Scientific names of organisms consist of a genus name and a species epithet, with the latter often reflecting aspects such as morphology, ecology, distribution, and cultural background. Traditionally, researchers have manually labeled species names by carefully examining taxonomic descriptions, a process that demands substantial time and effort when dealing with large datasets. This study evaluates the feasibility of automatic species name labeling using large language model (LLM) by leveraging their text classification and semantic extraction capabilities. Using the spider's name dataset compiled by Mammola et al., we compared LLM-based labeling results—enhanced through prompt engineering—with human annotations. The results indicate that LLM-based classification achieved high accuracy in Morphology, Geography, and People categories. However, classification accuracy was lower in Ecology & Behavior and Modern & Past Culture, revealing challenges in interpreting animal behavior and cultural contexts. Future research will focus on improving accuracy through optimized few-shot learning and retrieval-augmented generation techniques, while also expanding the applicability of LLM-based labeling to diverse biological taxa.**

***Keywords—Automated Labeling, Etymology Analysis, Taxonomic Nomenclature, Large Language Model, Natural Language Processing***

## I. INTRODUCTION

Humans have long sought to construct systematic classification methods to understand the complexity of natural phenomena and objects. These efforts serve as a foundation for uncovering patterns and interrelationships in nature, facilitating the accumulation of scientific knowledge. Various approaches have contributed to this goal, including Mendeleev's periodic table [1] and frameworks for categorizing human behavior and personality traits [2]. Such classifications attempt to function as universal tools for organizing the complexity of both nature and society. Among the most enduring scientific classification systems is the binomial nomenclature introduced by Carolus Linnaeus, which remains widely utilized today for identifying and comparing both living and fossil organisms [3]. In traditional taxonomy, scientific names have been devised based on multiple categories, including morphology, ecology, distribution, eponyms (human-derived species names), and cultural references. These etymological elements reflect the characteristics of organisms, the intentions of taxonomists, and the historical context in which species were named, thereby contributing to a deeper understanding of natural history. However, analyzing these etymologies requires meticulous examination of original descriptions and related literature. When dealing with large datasets, labeling species names may take several years. For instance, Mammola et al. [4] spent two years labeling approximately 48,000 spider names. Given the significant time investment required, a more efficient approach to classifying taxonomic names is highly desirable. In this context, advances in Natural Language Processing (NLP) present a promising avenue for improving efficiency. In particular, Large Language Model (LLM) has demonstrated substantial applicability across various domains, potentially influencing taxonomy as well.

Recent advancements in NLP have significantly enhanced the capabilities of LLM. These models, trained on vast textual datasets, extract statistical patterns and perform deep learning-based inference and text generation, achieving human-level or superior performance in tasks such as contextual understanding, semantic extraction, and text classification [5]. With the increasing commoditization of LLM, they have become accessible through APIs, enabling cost-effective integration into various applications, including automated labeling [6]. Compared to traditional manual approaches, LLM can significantly reduce time and labor while maintaining high accuracy, leading to growing expectations for automating taxonomic labeling. The application of LLM to etymological labeling could greatly enhance efficiency by leveraging their extensive linguistic knowledge and contextual understanding. Since Latin is commonly used in biological nomenclature but is no longer a widely spoken language, LLM may provide a particularly useful tool for interpreting species name origins. As taxonomy is a low-resource domain with limited explicit etymological data, LLMs can help infer origins through contextual reasoning. Unlike keyword searches or rule-based approaches, LLM can analyze entire texts, infer the intent of taxonomists, and structure information with higher accuracy. Consequently, automated etymology analysis could lead to significant improvements in taxonomic labeling efficiency while also capturing subtle nuances and cultural contexts that might otherwise be overlooked.

This study evaluates the feasibility of LLM-based automatic labeling in taxonomy, addressing the substantial costs and time constraints associated with manual classification of species name etymologies. Specifically, we compare the accuracy and efficiency of LLM-based labeling with traditional manual methods through quantitative and visual analysis. Our investigation focuses on three key aspects: i) Assessing the extent to which LLM can reduce labor costs while maintaining accuracy compared to traditional manual methods. ii) Examining how different prompt engineering techniques influence labeling accuracy and identifying optimal design strategies. iii) Comparing the LLM-based labeling results with human-based labels to quantify labeling accuracy. Through these evaluations, we demonstrate the potential of LLMs as a viable alternative for taxonomy-related labeling tasks. In particular, this study proposes a method for generating labels through Chain-of-Thought (CoT) inference using limited information such as morphological components of scientific names and the year of nomenclature, even in the absence of explicitly documented etymologies. This represents an early example of automated inference within low-resource domains like taxonomy. By leveraging taxonomy-specific prompt design and temporal information, we aim to uncover both the advanced applicability and current limitations of LLM-based labeling in specialized scientific contexts.

The primary contributions of this study are as follows:

i) LLM-based automatic labeling is demonstrated to be a viable alternative for taxonomic research by maintaining accuracy while significantly reducing processing time and human resource requirements, as evidenced by the comparison with manually labeled data.

ii) Prompt engineering techniques based on CoT are systematically evaluated to assess their impact on labeling accuracy, demonstrating their effectiveness in low-resource settings by highlighting how incorporating historical contextual information can improve performance.

iii) Quantitative and visual performance assessments using diverse evaluation metrics provide detailed insights into the accuracy and consistency of LLM-generated labels, offering a comprehensive evaluation of their applicability in taxonomic labeling.

The rest of this paper is structured as follows. Section II reviews background research on etymology analysis in taxonomy and LLM-based labeling. Section III describes the overall framework for LLM-based labeling and the prompt engineering techniques employed. Section IV presents the experimental setup and results, while Section V discusses the effectiveness and limitations of LLM-based labeling. Finally, Section VI concludes the study.

## II. Related Works

### A. *Etymology Analysis in Taxonomy*

Scientific names serve not only as a means of species identification but also as reflections of scientific tradition, cultural background, and creativity. Mammola et al. demonstrated that species naming in spiders is influenced not only by scientific criteria but also by personal dedications and cultural factors. Additionally, Macêdo et al. [7] analyzed rotifers and microcrustaceans, revealing a shift in species naming from scientifically grounded principles to geographical references and dedications to notable figures. Their statistical analysis highlighted a strong gender bias, showing that while male scientists' names were frequently used, those of female scientists remained significantly underrepresented. Vendetti [8] further confirmed this bias in mollusk eponyms, reporting that only 10.6% of species named after individuals honored female scientists.

Furthermore, the use of eponyms also raises ethical concerns. Guedes et al. [9] argued that eponyms can perpetuate colonialism and racial inequalities, deeming their usage inappropriate in 21st-century biological nomenclature. Conversely, Antonelli et al. [10] defended eponyms as a valuable means of recognizing scientific contributions, opposing any outright ban. A parallel debate exists regarding the incorporation of indigenous names into taxonomy. Wright and Gillman [11] criticized the dominance of Latin and Greek in existing taxonomic names, emphasizing that traditional indigenous knowledge is insufficiently represented. In response, Gillman & Wright [12] proposed clarifying standards for formally integrating indigenous names into scientific nomenclature, rather than altering existing naming rules. Another point of contention is whether ethical revisions of species names might compromise taxonomic stability. Ceríaco et al. [13] cautioned that renaming species for ethical reasons could disrupt the consistency and universality of scientific data, arguing that the International Commission on Zoological Nomenclature (ICZN) should not engage in ethical adjudication. In contrast, Jiménez-Mejías et al. [14] emphasized the need for stable and universal naming rules to ensure reliable knowledge sharing, advocating for careful evaluation before modifying species names.

The relationship between linguistic diversity and taxonomic naming is also a crucial issue, as English-centric renaming and the dominance of Latin and Greek are seen as barriers to inclusivity, potentially disadvantaging non-English-speaking researchers and restricting naming diversity [15, 16]. Moreover, the influence of species naming on research direction has also been explored, with findings indicating that taxa with plant-derived names are more frequently chosen for host-associated differentiation (HAD) studies, suggesting that species names themselves can shape research priorities [17, 18].

These discussions highlight that biological naming practices are shaped by a complex interplay of scientific tradition, gender bias, sociocultural impact, ethical considerations, and linguistic diversity. Nevertheless, addressing these challenges necessitates a comprehensive labeling effort for scientific names, a process that remains labor-intensive.

### B. *Automated Labeling Using LLM*

The advancement of LLM has expanded the potential for automated labeling across various NLP domains. Notably, improvements in zero-shot and few-shot learning have enabled models to perform tasks with minimal explicit training examples or prior knowledge [19]. Brown et al. [20], in an early study on few-shot learning, demonstrated that GPT-3 could generalize to novel tasks with only a small number of examples, highlighting how large-scale pretraining endows LLM with broad knowledge and adaptability. Chamieh et al. [21] later examined LLM performance in zero-shot and few-shot answer evaluation, confirming their effectiveness in simple classification tasks. Zhao et al. [22] further observed that LLM exhibits robust multilingual performance, attributing this capability to an internal translation mechanism that converts input into English representations before processing.

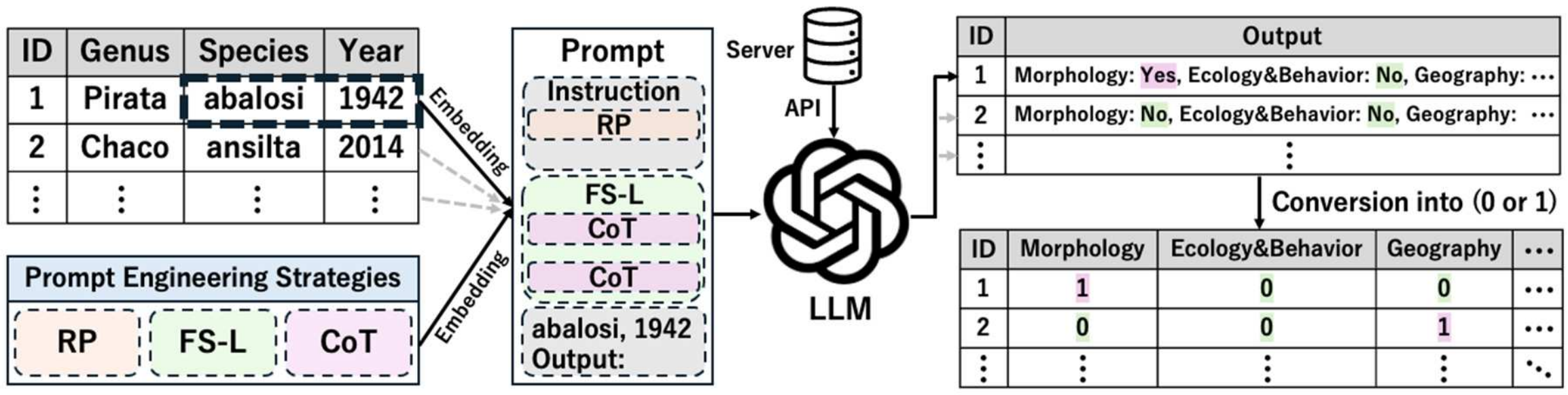

Fig. 1. Experiment Workflow for LLM-Based Automated Labeling.

Despite these advances, limitations remain in zero-shot learning, and few-shot learning requires careful prompt engineering for optimal performance. Kojima et al. [23] proposed improving zero-shot performance by incorporating the phrase "Let's think step by step" into prompts. Perez et al. [24] introduced standardized evaluation benchmarks to better assess few-shot generalization capabilities, while Wei et al. [25] demonstrated that CoT prompting significantly enhances accuracy in complex reasoning tasks. Additionally, Kong et al. [26] showed that incorporating role-playing techniques into prompts further improves LLM performance. Given these findings, Li [27] emphasized that prompt design significantly influences LLM outputs, arguing that well-structured instructions and examples are critical for successful few-shot labeling. In this study, we compare multiple prompt engineering strategies to identify the most effective taxonomy-specific prompt design.

## III. Experiment Setup

### A. Dataset

For this study, we utilize the JAFList dataset [28] compiled by Mammola et al., which systematically organizes spider scientific names along with information on their distribution and discovery years. A key feature of this dataset is that etymology-based categorization has been manually conducted through literature review, with species epithets (Species) labeled according to the categories shown in Table I. As a result, this dataset serves as ground truth for evaluating the effectiveness of LLM-based automatic labeling. By comparing LLM-based labels with human-based labels, we can assess the feasibility and accuracy of LLM-based taxonomic labeling. The dataset consists of 48,464 entries, with each species classified into one or more categories. Notably, since some Species belong to multiple categories, the sum of category counts does not equal the total number of data points.

### B. Experiment Workflow

To evaluate the effectiveness of LLM-based automatic labeling, this study utilizes a manually labeled dataset as the ground truth, comparing it with LLM-generated labels. The overall process follows a structured workflow, as illustrated in Fig. 1, beginning with the extraction of the Species from the full scientific name. The extracted Species is then embedded within a carefully designed prompt that specifies the expected output format, ensuring that the model can classify it into predefined categories accurately. Once the prompt is prepared, it is provided as input to the LLM, which performs inference based on the given Species. To maintain consistency in the results, the temperature parameter is set to 0, ensuring that the model produces deterministic outputs rather than relying on probabilistic variations. This setting allows the LLM to always select the most confident inference, reducing randomness in labeling outcomes.

TABLE I. Categories for Taxonomic Labeling

| Category Name | Description |
|---|---|
| **Morphology** | Naming based on morphological characteristics such as the size, body shape, and color |
| **Ecology and Behavior** | Naming based on habitat and behavioral traits |
| **Geography** | Naming based on species distribution or the location where they were discovered |
| **People** | Naming after scientists or other notable individuals |
| **Modern and Past Culture** | Naming derived from mythology, music, movies, pop culture, and other cultural references |
| **Other** | Naming that does not fit into the above categories |

For this experiment, GPT-4o-mini was selected as the LLM, accessed via OpenAI's API. Although this model has lower accuracy compared to the more advanced o1 and o3 models, it was chosen for its low cost and high-speed inference capabilities. Since taxonomic labeling does not require highly sophisticated reasoning, the trade-off in accuracy was deemed acceptable for this study. Following inference, the LLM-generated outputs, initially in natural language, undergo post-processing to conform to a structured format. The prompt was designed to enforce a clear output structure, allowing easy conversion of responses into binary labels (0 or 1) for each category. These results are then stored in a CSV file for subsequent analysis. All experiments were conducted over a four-day period, from February 28, 2025, to March 3, 2025.

### C. Prompt Design

For LLM-based automatic labeling, Species must be presented within a structured prompt. In this study, we designed a prompt incorporating three prompt engineering techniques to enhance labeling accuracy, as shown in Fig. 2.

- Role-Playing (RP): Assigning a specific role to the LLM to elicit appropriate outputs. This technique helps generate specialized and consistent responses.

TABLE II. ACCURACY COMPARISON OF PROMPT ENGINEERING TECHNIQUES FOR AUTOMATED LABELING

| Prompt | Morphology | Ecology & Behavior | Geography | People | Modern & Past Culture | Other | Average |
|---|---|---|---|---|---|---|---|
| **Full** | **93.0%** | 95.3% | 90.1% | 93.7% | **95.7%** | **91.6%** | **93.2%** |
| **w/o Year** | **93.0%** | 95.0% | 89.0% | **94.0%** | 92.0% | 90.0% | 92.2% |
| **RP** | 92.2% | 95.2% | 88.0% | 90.7% | 95.4% | 87.9% | 91.6% |
| **FS-L** | 92.2% | **95.8%** | **90.8%** | 93.7% | 92.2% | 90.6% | 92.6% |
| **CoT** | 91.0% | 93.3% | 86.0% | 89.6% | 93.8% | 82.1% | 89.3% |

***You are an expert in taxonomic nomenclature.*** *Scientific species epithets are derived primarily from Latin or Latinized Greek, and they often reflect morphological, ecological, geographical, or cultural aspects. Please analyze the following species epithet (do not include the genus) and assign it to one or more of the following six categories:*

1. *Morphology: referring to physical appearance, form, size, color, etc.*
2. *Ecology & Behavior: referring to habitat, behavior, or ecological niche.*
3. *Geography: referring to the region, country, or type locality.*
4. *People: dedicated to a person (e.g., scientist or notable individual).*
5. *Modern & Past Culture: referring to cultural references, modern or historical.*
6. *Other: if the epithet has a meaning but does not clearly fall into any of the above categories, or if it is a nonsensical name.*

*Internally, perform a detailed chain-of-thought analysis explaining how you arrived at your classification. However, in your final output, provide only the final answer in the following format (do not include your internal reasoning):*
***Format****: Morphology: [Yes/None], Ecology & Behavior: [Yes/None], Geography: [Yes/None], People: [Eponym (male)/Eponym (female)/None], Modern & Past Culture: [Yes/None], Other: [Yes/None]*
*The species was described in the year, which may influence whether it references an older historical figure, a more modern cultural reference, or a classical Latin/Greek etymology. Consider this date in your classification.*
***Few-shot examples****:*
*Example 1*
*Species epithet: jaculator*
***Year****: 1910*
***Chain-of-Thought****: (omitted)*
*Final Answer: Morphology: None, Ecology & Behavior: Yes, Geography: None, People: None, Modern & Past Culture: None, Other: None*
*Example 2 (omitted)*
*Example 3 (omitted)*
*Example 4 (omitted)*
*Now, please analyze the following species epithet and output only the final answer.*
***Species epithet: {Species}***
***Year: {Year}***
***Final Answer:***

Fig. 2. Prompt Design for Automated Taxonomic Labeling.

- Few-Shot Learning (FS-L): By including a few examples input-output pairs within the prompt, the model can recognize labeling patterns without requiring fine-tuning.
- CoT: Structuring the prompt to encourage step-by-step reasoning improves accuracy in complex labeling tasks by making the inference process explicit. These techniques are particularly beneficial for taxonomic etymology labeling, where labeled training data is limited.

Additionally, we incorporated the year of species description as a temporal feature, allowing the LLM to consider historical and cultural contexts in labeling. This prompt was optimized through preliminary experiments to ensure high efficiency in automated etymology labeling. To maintain consistency in output format, the LLM was instructed to return binary responses (Yes or None) for each category. This facilitated straightforward post-processing, where the responses were converted into 1 or 0 labels for structured data analysis. By implementing these prompt engineering strategies, this study aims to maximize the accuracy of LLM-based automatic labeling in taxonomy.

## IV. EXPERIMENT AND ANALYSIS

### A. Experiment Design

To evaluate the effectiveness of LLM-based automatic labeling, we conducted a four-stage experiment comparing LLM-generated labels with human-annotated labels. In the first stage, we optimized the prompt engineering techniques to enhance the accuracy of LLM-based labeling. This involved evaluating RP, FS-L, and CoT reasoning both individually and in combination. Additionally, we compared results with and without temporal information (year of species description) to determine its impact. The second stage focused on analyzing category-wise similarity between LLM-generated labels and human annotations. By performing a quantitative and visual comparison, we assessed the consistency and accuracy of LLM-based labeling for each category. In the third stage, we investigated temporal patterns using a Generalized Additive Model (GAM) to compare time-series trends across different categories. This analysis enabled a more detailed assessment of how well the LLM captured temporal etymological patterns. Finally, in the fourth stage, we examined spatial patterns by comparing results across different geographic regions. This evaluation helped determine whether the LLM's labeling performance varied depending on regional factors. Through these four stages, we systematically assessed the effectiveness of LLM-based automatic labeling for taxonomic etymology classification.

### B. Evaluation of Prompt Engineering Techniques

To improve the accuracy of LLM-based labeling, we conducted an evaluation of different prompt engineering techniques and optimized prompt design accordingly. A test dataset of 1,000 randomly extracted samples was used to compare labeling accuracy under five conditions: RP, FS-L, CoT reasoning, a combination of all three techniques without Year information (w/o Year), and a full prompt including Year information (Full). The results are summarized in Table II, where each value represents the accuracy for a given category, along with the average accuracy across all categories.

The findings indicate that the Full achieved the highest overall accuracy. Although certain categories exhibited slightly lower accuracy compared to other approaches, the differences were minor and within the range of statistical error. Comparing the Full with w/o Year, we observed that the Full outperformed the latter in nearly all categories, suggesting that Year information plays a crucial role in LLM-based etymology labeling. The Modern & Past Culture category, in particular, showed a notable improvement when Year information was included, indicating that the LLM effectively associates temporal context with cultural references. Among the individual prompt engineering techniques, FS-L had the strongest impact, suggesting that this approach is particularly effective for species name labeling.

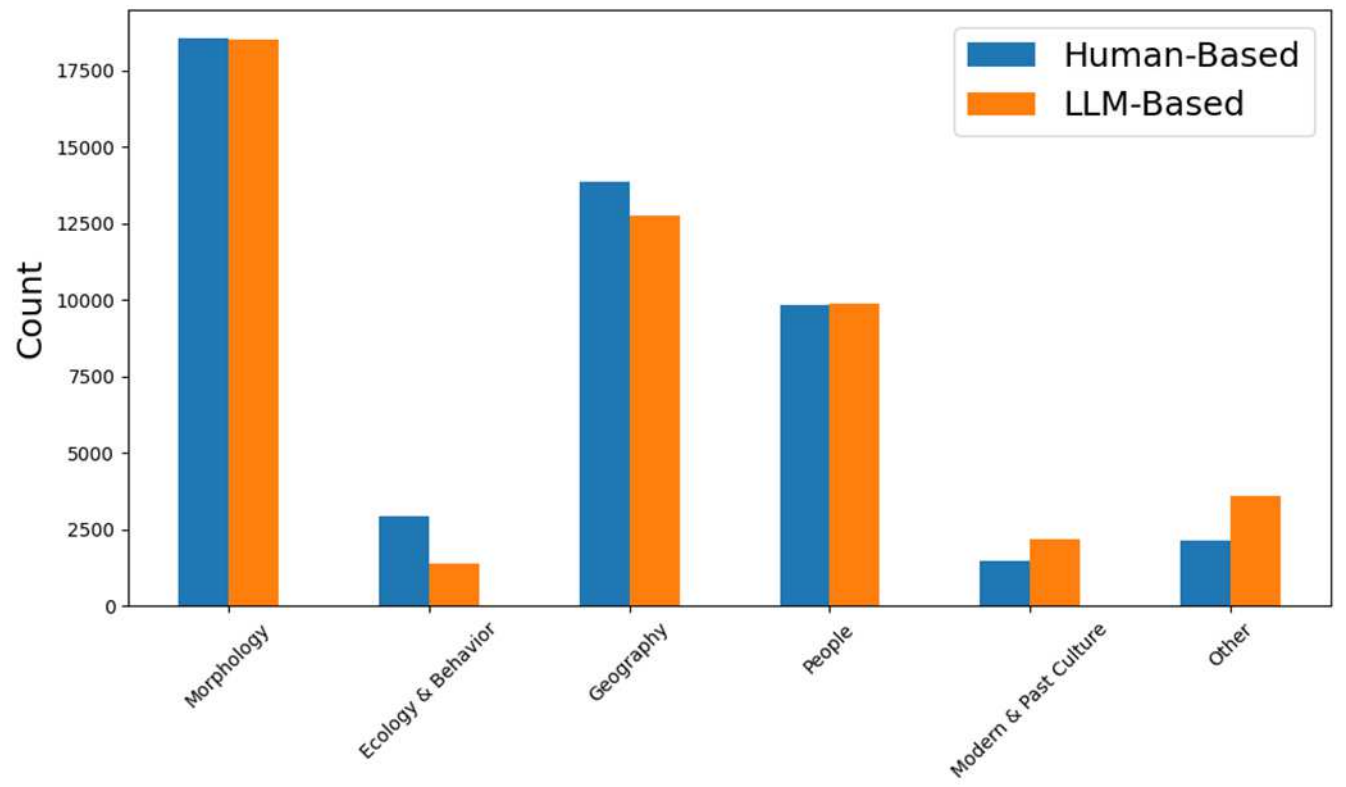

Fig. 3. Comparison of LLM and Human Labeling Patterns.

TABLE III. PERFORMANCE OF LLM-BASED LABELING

| Category | Accuracy | Precision | Recall | F1 |
|---|---|---|---|---|
| **Morphology** | 91.8% | 89.6% | 89.5% | 89.5% |
| **Ecology & Behavior** | 94.6% | 63.8% | 29.7% | 40.5% |
| **Geography** | 88.9% | 83.7% | 77.2% | 80.3% |
| **People** | 94.1% | 85.8% | 86.0% | 85.9% |
| **Modern & Past Culture** | 94.7% | 26.9% | 40.0% | 32.2% |
| **Other** | 91.2% | 21.8% | 36.8% | 27.4% |

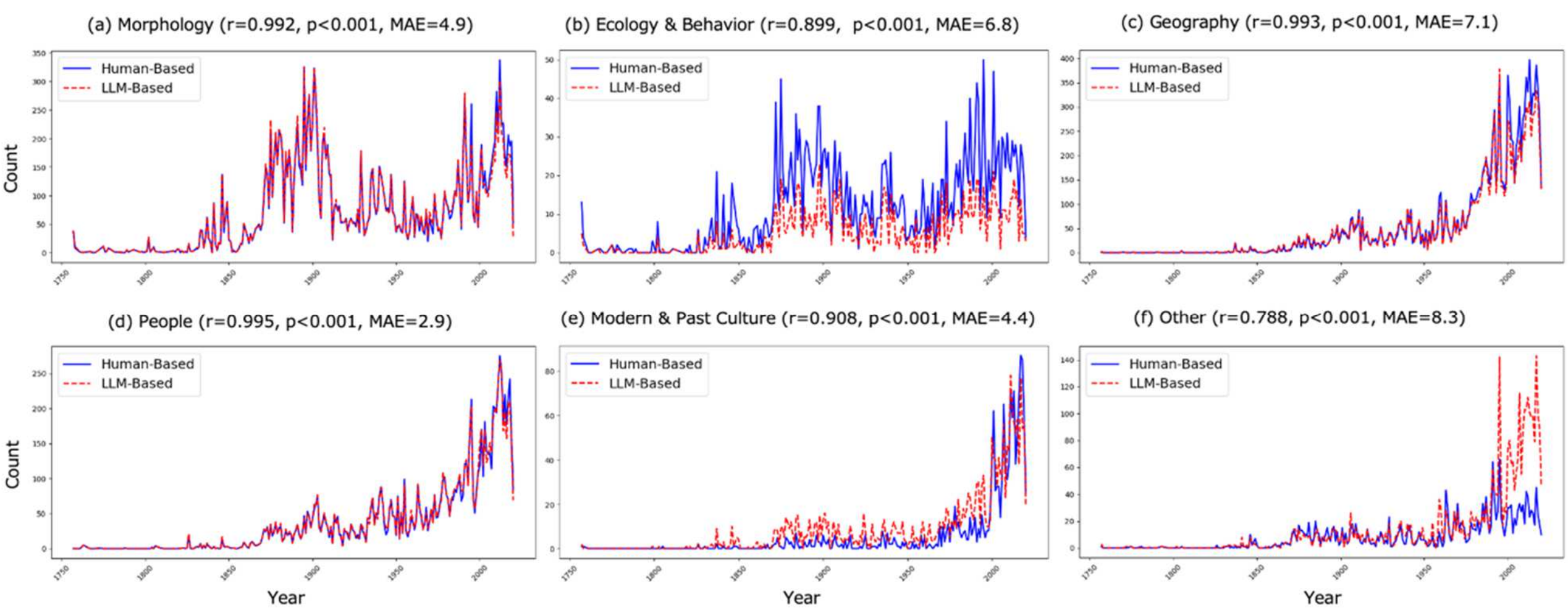

Fig. 4. Comparison of Time-Series Trends in LLM and Human Labeling.

## *C. Evaluation of Basic Patterns*

To investigate the similarity between LLM-based labeling and human-based labeling, we compared labeling patterns across different categories. The results of this comparison are shown in Fig. 3.

The LLM-based automatic labeling process was completed within a single day, even when incorporating idle time to avoid API rate limits. This represents a substantial reduction in processing time compared to the two-year human-labeling effort documented in previous studies. In terms of accuracy, LLM-generated labels closely matched human-based labels in categories such as Morphology and People. However, slight discrepancies were observed in other categories, where Ecology & Behavior and Geography were occasionally misclassified into Modern & Past Culture or Other.

Table III presents the performance metrics for each category. The results indicate that accuracy was generally high across all categories, confirming that the LLM effectively identified non-relevant (None) cases. Further analysis of precision and recall revealed that Morphology, Geography, and People achieved relatively high performance, suggesting that LLM-based labeling was closely aligned with human annotations for these categories. On the other hand, recall scores were lower for some categories, indicating that the LLM struggled to fully reproduce the human-labeled dataset. In particular, the Modern & Past Culture category exhibited the lowest recall except for the Other category, suggesting that LLM may find it more challenging to classify etymological references related to cultural and historical contexts.

## *D. Evaluation of Temporal Patterns*

To assess the temporal trends in LLM-based labeling, we analyzed time-series patterns across categories. Fig. 4 presents a time-series plot of each etymology category alongside the corresponding LLM-labeled results. The Pearson correlation coefficient analysis indicates that the temporal trends of all categories, except Other (Fig. 4f), annotations. While Ecology & Behavior shows a lower number of instances, it still exhibits a strong overall agreement between LLM-based and human number of predicted instances overall, the temporal trends are well captured (Fig. 4b). A more detailed evaluation based on the Mean Absolute Error (MAE) reveals that the People category shows an average discrepancy of only 2.9 mislabeled instances per year, demonstrating the high reliability of LLM-based labeling (Fig. 4d). In contrast, Geography has an average of 7.1 mislabeled instances per year. However, considering the total dataset size, this still translates to an overall accuracy of approximately 90%, indicating a high level of performance (Fig. 4c).

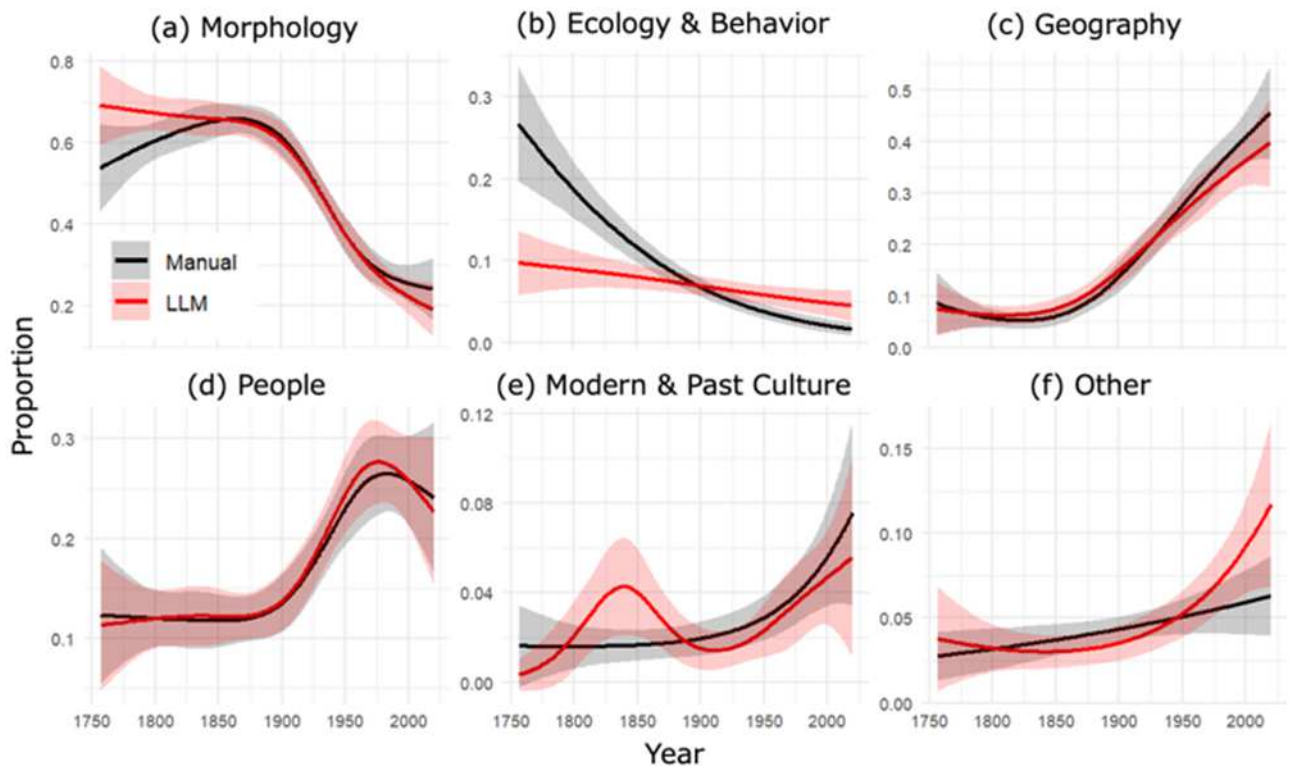

Fig. 5. Comparison of GAM-Based Temporal Trends in LLM and Human Labeling.

Next, we performed a visual comparison using GAM to estimate the relative proportions of each etymology category over time. The results, shown in Fig. 5, illustrate smoothed trends with 95% confidence intervals. The model successfully reconstructs the temporal patterns of categories such as Morphology (Fig. 5a), Geography (Fig. 5c), and People (Fig. 5d) with high accuracy. Although Modern & Past Culture exhibits some deviations, the majority of the pattern aligns well with the ground truth (Fig. 5e). However, Ecology & Behavior remains a challenging category for LLM-based labeling (Fig. 5b).

We further visualized the annual fluctuations in the proportion of each etymology category and compared them to the GAM-predicted trends, as shown in Fig. 6. Here, the 95% confidence intervals indicate that the overall data distribution remains similar across both human and LLM-based labels. These findings suggest that LLM-based automatic labeling is highly effective for Morphology, Geography, and People categories, moderately effective for Modern & Past Culture, and less reliable for Ecology & Behavior.

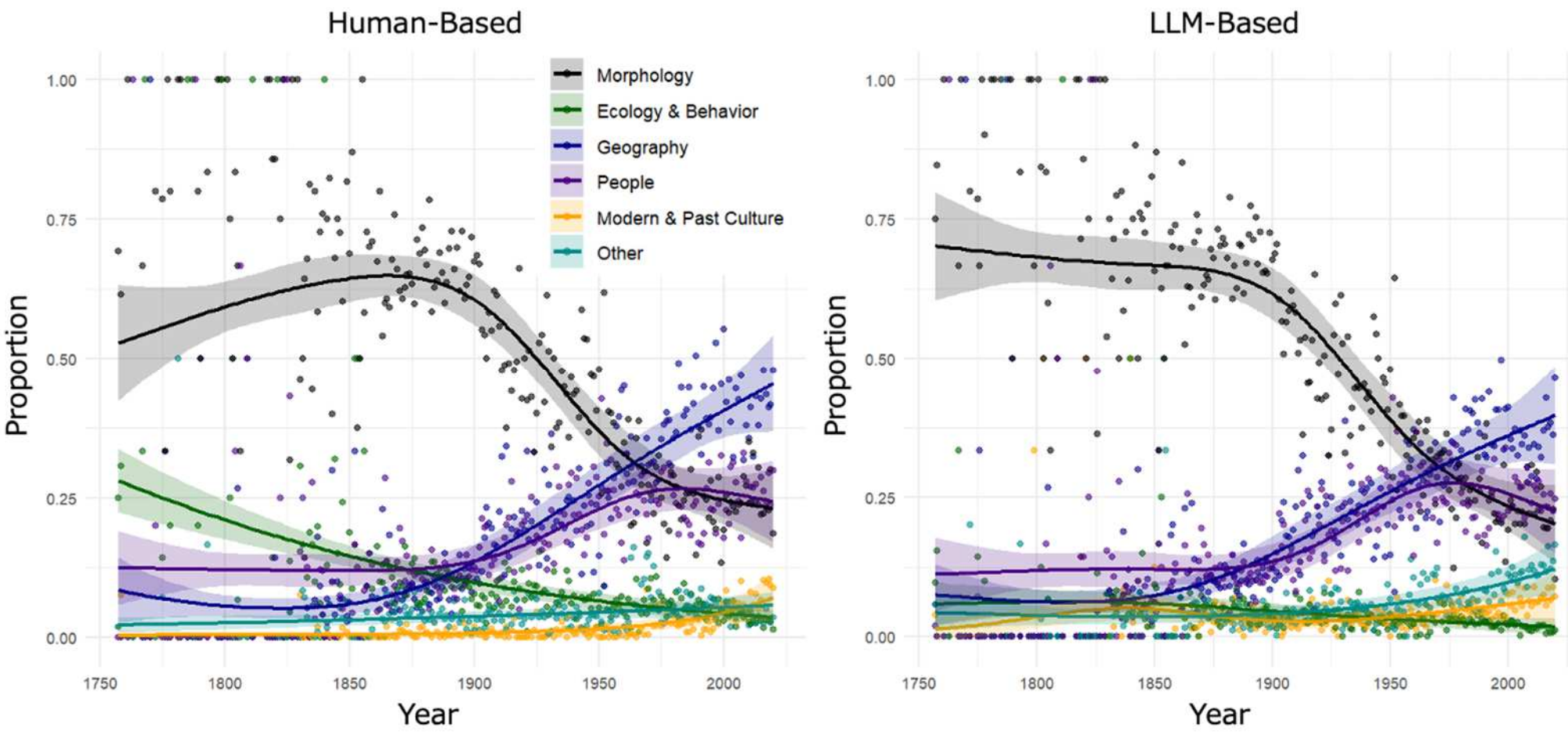

Fig. 6. Comparison of Annual Fluctuations and GAM Predictions in LLM and Human Labeling.

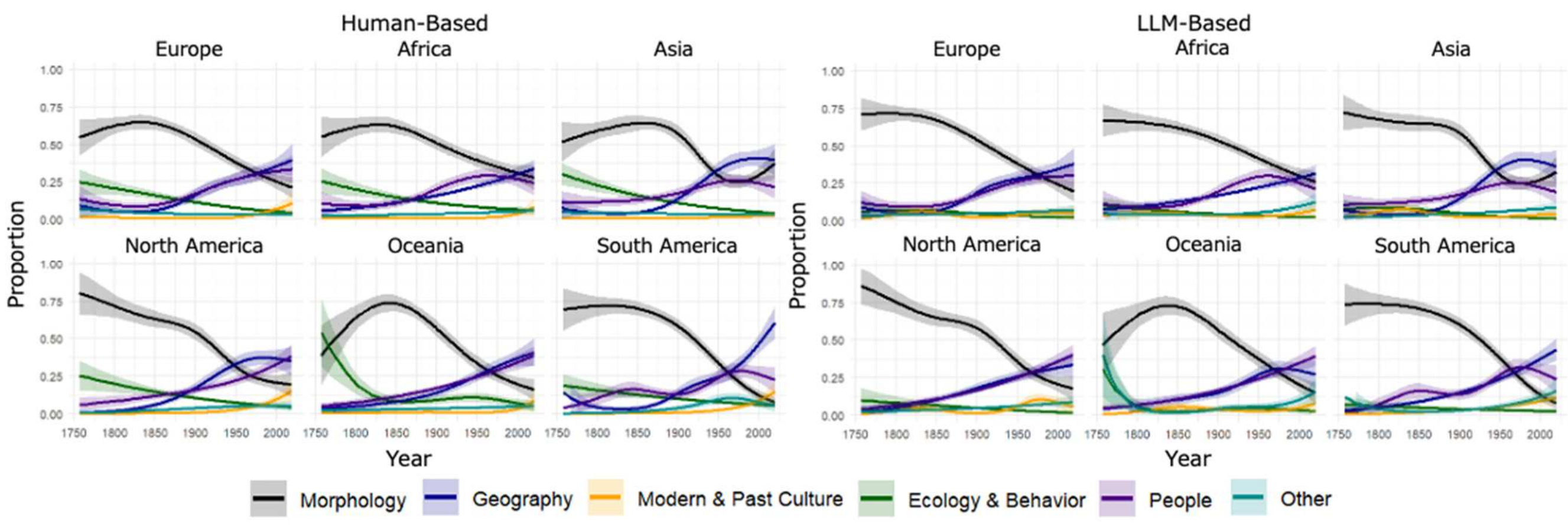

Fig. 7. Comparison of Temporal Trends in LLM and Human Labeling Across Regions.

## *E. Evaluation of Spatial Patterns*

To examine whether LLM-based labeling performance varies across different geographic regions, we analyzed the category distribution by region. Using GAM, we estimated the temporal changes in relative category proportions across different regions and visualized the results using smoothed functions, as shown in Fig. 7. The results demonstrate that LLM effectively predicts regional trends, with particularly high consistency in Oceania and South America. Overall, no significant discrepancies in labeling accuracy were observed across regions. However, species described in earlier historical periods exhibited lower similarity between LLM-based labels and human-based labels. This effect was particularly pronounced in the Ecology & Behavior category,

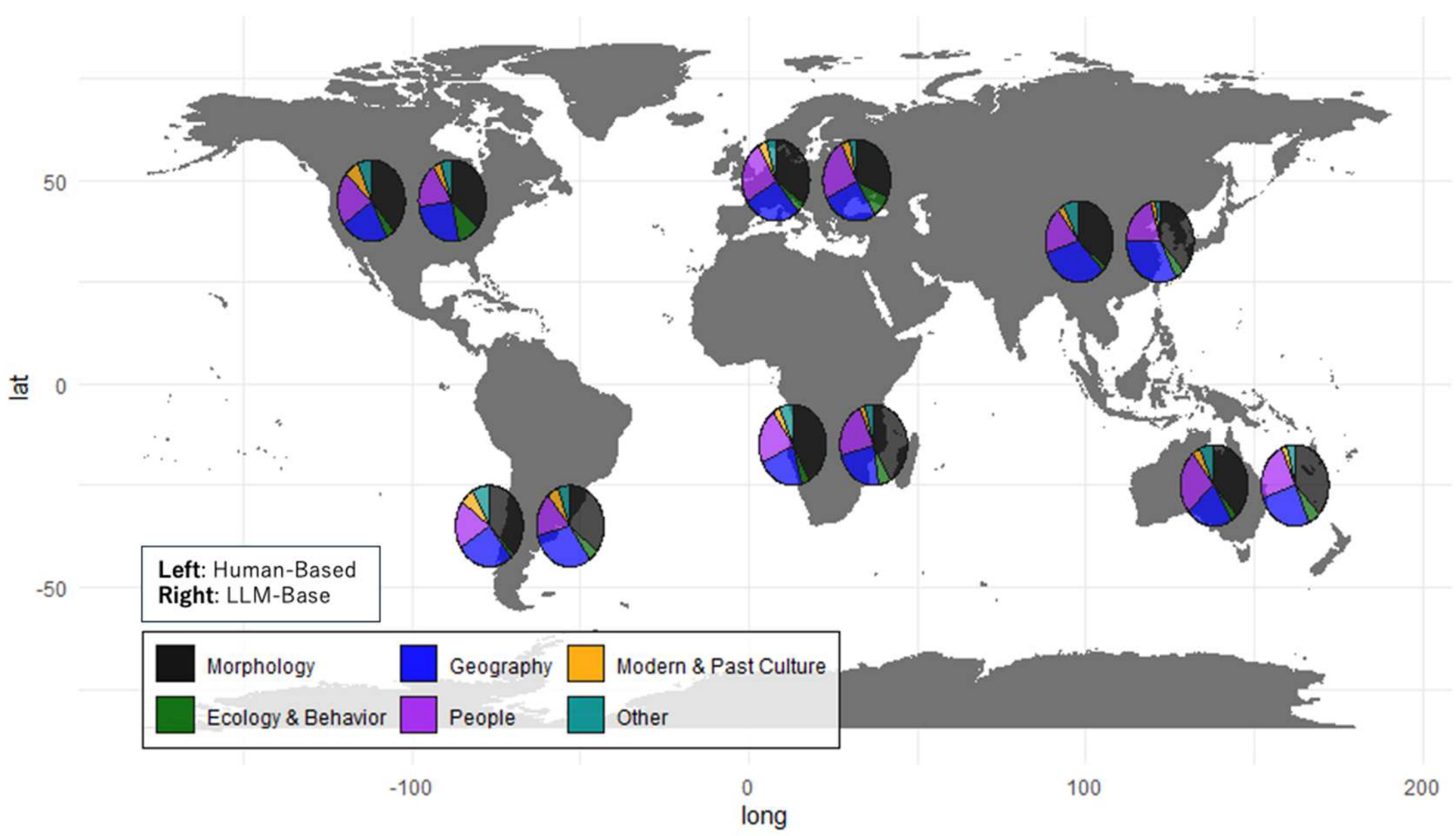

Fig. 8. Comparison of Regional Distributions of Etymology Categories in LLM and Human Labeling.

where LLM-based labeling showed both underestimation and overestimation biases across all regions.

Fig. 8 presents the proportion of each category across different regions. The Ecology & Behavior category was found to be labeled at a slightly higher rate by the LLM, but apart from this, no major regional discrepancies were observed. Overall, these results indicate that LLM-based labeling exhibits strong regional consistency and a high degree of similarity to human-based labeling.

## V. Discussion

This study examined the effectiveness and limitations of LLM-based automatic labeling for species name etymology in low-resource settings by comparing LLM-generated labels with human-based labels. The results demonstrated that a task requiring two years with manual labeling could be completed within a single day using LLM-based labeling. Additionally, high labeling accuracy was observed in Morphology, People, and Geography, achieving near-human performance. Since these three categories account for a significant portion of etymology classifications, the utility of LLM-based labeling is strongly emphasized. On the other hand, Modern & Past Culture and Ecology & Behavior exhibited lower labeling accuracy, with the discrepancy becoming more pronounced for older taxonomic names in the Ecology & Behavior category. This trend suggests that while LLM possesses extensive knowledge about morphology, geography, and personal names, they have limited understanding of cultural backgrounds and ecological behaviors. In the Modern & Past Culture category, the primary challenge is that LLM training data is predominantly derived from modern web-based information, making it difficult to accurately capture historical cultural contexts. The gap between contemporary knowledge and the cultural landscape at the time of species naming likely contributed to reduced labeling accuracy. For Ecology & Behavior, the lower accuracy may be attributed to the limited availability of ecological information within LLM training data, restricting its ability to correctly interpret Species with ecological significance.

However, one critical aspect to consider is the strong potential for generalizing LLM-based labeling across various biological taxa. While this study focused on spider taxonomy, the results suggest that LLMs can be effectively applied to other groups, including vertebrate, which exhibit distinct etymological patterns. Given LLMs' high accuracy in morphology- and geography-based classifications, they are likely well-suited for broader taxonomic applications, including fossil taxa and microorganisms, which pose challenges due to historical naming conventions and limited documentation. Additionally, as some taxa have scientific names shaped by regional linguistic traditions, LLMs' ability to process multilingual and context-dependent data suggests they could bridge nomenclatural gaps across languages.

This study also has several limitations. The ground truth dataset used for evaluation was constructed through manual literature review, meaning that not all labels were perfectly validated, and some annotations contain errors. As a result, even when LLM labeling was correct, discrepancies with human-labeled data might have led to the underestimation of accuracy. Additionally, the possibility exists that LLM training data contained part of the dataset used in this study. However, since the primary goal was to evaluate LLM's understanding of species names rather than their ability to infer unknown data, potential data leakage does not significantly impact the validity of our conclusions.

Overall, the results of this study highlight the importance of collaboration between taxonomists and LLM specialists. Such cooperation can streamline labeling in low-resource environments and accelerate new discoveries in taxonomy. Future research should focus on enhancing the classification accuracy of Modern & Past Culture and Ecology & Behavior, with potential strategies including expanding FS-L samples, integrating Retrieval-Augmented Generation (RAG) techniques, and incorporating human feedback, while also evaluating cross-taxonomic consistency to ensure that LLM-based labeling remains robust and adaptable across diverse biological groups.

## VI. CONCLUSION

Classifying species name origins is a vital but time-intensive task, especially in low-resource taxonomic domains. This study evaluated LLM-based automatic labeling by comparing prompt-engineered outputs with human annotations. Results showed that LLMs significantly reduce processing time while maintaining high accuracy in major categories such as Morphology, Geography, and People. Prompt engineering techniques—including CoT reasoning, FS-L, and temporal information—proved especially effective in improving accuracy for complex categories, highlighting their value in settings with limited explicit data. Quantitative and visual evaluations further revealed areas of strength and weakness, with lower performance in Ecology & Behavior and Modern & Past Culture due to challenges in historical and ecological understanding.

Future research should aim to improve accuracy through enhanced FS-L, temporal data integration, and human feedback, while validating the approach across diverse taxa. This study highlights LLM's potential as a powerful tool for advancing automated species name labeling in taxonomy.

## ACKNOWLEDGMENT

This work was supported by JST SPRING.